\documentclass{llncs}

\usepackage{mathptmx}
\usepackage{times}

\usepackage{multicol}
\usepackage[bookmarks=true]{hyperref}
\usepackage{amssymb}
\usepackage{float}
\usepackage{makeidx}
\usepackage{amsmath}
\usepackage{graphicx}
\usepackage{epsfig}
\usepackage{epsf}
\usepackage{psfrag}
\usepackage{verbatim}
\usepackage{lscape}
\usepackage{array}
\usepackage{subfigure}
\usepackage{wrapfig}
\usepackage{rotating}
\usepackage{wrapfig}
\usepackage{multirow}
\usepackage{undertilde}
\usepackage{mathrsfs}
\usepackage[mathcal]{euscript}
\usepackage[all]{xy}
\usepackage{color}
\usepackage{multirow}
\usepackage{array}
\usepackage{arydshln}
\usepackage{chngpage}
\usepackage{lipsum}
\usepackage{url}

\def\react{{\sc ReAct!}}

\def\ar{\leftarrow} 
\def\ii#1{\hbox{\it #1\/}}  
\def\beq{\begin{equation}}
\def\eeq#1{\label{#1}\end{equation}}
\def\ba{\begin{array}}
\def\ea{\end{array}}

\def\ccalc{{\sc CCalc}}
\def\cplus{${\cal C}+$}

\def\openrave{{\sc OpenRave}}

\def\chaff{{\sc Chaff}}
\def\minisat{{\sc MiniSAT}}
\def\manysat{{\sc ManySAT}}
\def\cplusasp{{\sc cplus2asp}}

\newcommand{\static}[2]{\ensuremath{{{\textrm{\bf caused\ }}#1{\textrm{\bf\ if\ }}#2}}}

\newcommand{\causes}[2]{\ensuremath{{#1\ {\textrm{\bf causes\ }}#2 }}}

\newcommand{\nonexecutableIf}[2]{\ensuremath{{\textrm{\bf nonexecutable\ }}#1  {\textrm{\bf\ if\ }} #2}}

\def\clasp{{\sc Clasp}\,}

\def\ba{\begin{array}}
\def\ea{\end{array}}
\def\beq{\begin{equation}}
\def\eeq#1{\label{#1}\end{equation}}

\def\ii#1{\hbox{\it #1\/}}

\begin{document}

\title{\react\  An Interactive Tool for\\ Hybrid Planning in Robotics}
\author
         {Zeynep Dogmus \and Esra Erdem \and Volkan Patoglu\\
}

\institute{Sabanc\i\ University, \.Istanbul, Turkey\\
\email{\{zeynepdogmus,esraerdem,vpatoglu\}@sabanciuniv.net}
}
\label{firstpage}

\maketitle

\begin{abstract}
We present \react, an interactive tool for high-level reasoning for cognitive
robotic applications. \react\ enables robotic researchers to describe robots' actions
and change in dynamic domains, without having to know about the
syntactic and semantic details of the underlying formalism in
advance, and solve planning problems using state-of-the-art
automated reasoners, without having to learn about their input/output language
or usage.
In particular, \react\ can be used to represent sophisticated
dynamic domains that feature concurrency, indirect effects of
actions, and state/transition constraints. It allows for
embedding externally defined calculations (e.g., checking for
collision-free continuous trajectories) into representations
of hybrid domains that require a tight integration of (discrete)
high-level reasoning with (continuous) geometric reasoning. \react\
also enables users to solve planning problems that involve
complex goals. Such variety of utilities are useful for robotic researchers
to work on interesting and challenging domains, ranging from service
robotics to cognitive factories.
\react\ provides sample formalizations of some action domains
(e.g., multi-agent path planning, Tower of Hanoi), as well as
dynamic simulations of plans computed by a state-of-the-art automated
reasoner (e.g., a SAT solver or an ASP solver).
  \end{abstract}


\section{Introduction} \label{Sec:Intro}

As the robotics technology makes its transition from repetitive
tasks in highly-structured industrial settings to loosely defined
tasks in unstructured human environments, substantial new challenges
are encountered. For instance, in order to be able to deploy robotic
assistants in our society, these systems are expected to robustly
deal with high complexity and wide variability of their surroundings
to perform typical everyday tasks without sacrificing safety.
Moreover, whenever more than one robotic agent is available in a
domain, these robots are expected to collaborate with each other
intelligently to share common tasks/resources. The complexity of the
tasks and the variability of environment place high demands on the
robots' intelligence and autonomy. Consequently, there exists a
pressing need to furnish robotic systems with high-level cognitive
capabilities.

The multidisciplinary field of robotics is diverse and the technical
background of robotics researchers are highly heterogenous. Even
though artificial intelligence (AI) planning and reasoning about
actions and change have been studied for decades in the field of
computer science, leading to various action description languages,
computational methods and automated reasoners, utilization of these
outcomes in robotic systems has only recently gained momentum,
thanks to increasing demands from new challenging domains, like
service robotics applications. However, as many of the robotics
researchers trained in diverse engineering aspects of robotics are
not familiar with these logic-based formalisms, underlying
theoretical AI concepts, and the state-of-the-art automated
reasoners, it is still a challenge to integrate high-level automated
reasoning methods in robotics applications.

In this paper, we introduce an interactive tool, called \react, to
fulfill this need in robotics. With \react, robotics researchers can
describe robots' actions and change in a dynamic domain, without
having to know about the syntactic and semantic details of the
underlying formalism in advance. Such dynamic domains may be quite
sophisticated, allowing concurrency, indirect effects of actions,
and state/transition constraints. They can also solve planning
problems, which may involve temporal complex goals, using a
state-of-the-art automated reasoner, without having to know about its
input/output language or usage. Furthermore, while computing a
(discrete) plan, some geometric constraints on continuous
trajectories can be embedded in domain description, to ensure
feasible (collision-free) plans.

\react\ utilizes two sorts of automated reasoners: SAT solvers
(e.g., \chaff~\cite{MoskewiczMZZM01}, \minisat~\cite{EenS03},
\manysat~\cite{HamadiJS09}) and ASP solvers (e.g.,
\clasp~\cite{gebserKNS07}). According to SAT~\cite{gomesKSS08}, the
idea is to formalize (in general NP-hard) computational problems as
a set of formulas in propositional logic so that the models of this
set of formulas characterize the solutions of the given problem;
and compute the models using SAT solvers. According to
ASP~\cite{lifschitz08,BrewkaET11}, the idea is similar; though the
problems are represented in a different logical formalism where
the formulas (called rules) look different, have a non-monotonic meaning and the models (called
answer sets~\cite{gelfondL91}) are computed using ASP solvers. Both
SAT and ASP have been applied in various real-world applications. For
instance, SAT has been used for software and hardware
verification~\cite{VelevB03} and planning~\cite{KautzS92}; ASP has
been applied to biomedical informatics \cite{ErdemEEO11}, space
shuttle
control~\cite{decision_support_systems:Nogueira00ana-prolog},
workforce management~\cite{RiccaGAMLIL2O12}, etc.

Although both SAT and ASP provide efficient solvers and their
formalisms are general enough to represent various kinds of
computational problems, their formalisms do not provide special
structures for a systematic formalization of dynamic domains. To
facilitate the use of such general knowledge representation
formalisms for representing dynamic domains, some other form of
logic-based formalisms, called action description
languages~\cite{gel98}, have been introduced. Further, to be able to
use SAT/ASP solvers for reasoning about actions and change, sound
transformations from action description languages to SAT and ASP
have been developed~\cite{McCainT98,LifschitzT99,GiunchigliaLLMT04}.

\react\ provides an interactive interface to systematically
represent actions and change in the action language
\cplus~\cite{GiunchigliaLLMT04}. 
It also allows the users to solve planning problems using state-of-the-art SAT/ASP solvers. 
In that sense, \react\ helps robotic
researchers to learn, understand, and use high-level representation
and reasoning concepts, formalisms, methods, and reasoners.

\section{Reasoning about Actions and Change} \label{Sec:Reasoning}

For an agent to act intelligently in a dynamic domain, one of the
essential high level cognitive functions for that agent is
reasoning about its actions and the change that is directly or
indirectly caused by these actions. For instance, AI planning is one
of such reasoning tasks: a robotic agent should be able to
autonomously find a sequence of actions to execute, in order to
reach a given goal from the initial state she is at. To perform
reasoning tasks, an agent should know about which actions she can
execute, as well as how these actions change the world. For that, we
can describe the actions of the agent in a logic-based formalism so
that the agent can autonomously perform reasoning tasks (e.g., find
plans) by using an automated reasoner based on  logic-based
inferences.

On the other hand, representing actions of an agent in a logic-based
formalism requires some background in logic as well as the specific
representation language. Consider, for instance, a mobile robot that
can go from one location to another location, as well as pick and
place some boxes in these locations. States of the world can be
described by means of three fluents (i.e., predicates whose truth
value may change over time): one describing the location of the
robot, one describing the locations of objects, another describing
whether the robot is holding an object or not. Then, we describe the
action of a robot going to a location $y$, by representing the
preconditions (i.e., the robot is not at $y$) and the direct effects
(i.e., the robot is at $y$). We also describe the indirect effects
of actions, and state/transition constraints, like:
\begin{itemize}
\item[]\emph{Ramifications:} If the robot is holding a box $b$, and the
robot goes to some location $y$, then as an indirect effect of this
action the location of the box $b$ becomes $y$ as well.\\

\item[]\emph{State Constraints:} Every robot (or box) cannot be at two different
locations, and two boxes cannot be at the same location at any state
of the world.
\end{itemize}
Further, we need the \emph{commonsense law of inertia:} if an action
does not change the location of the robot (resp. a box), then the
robot's (resp. a box's) location remains to be the same.

Figures~\ref{fig:move-sat}, \ref{fig:move-asp},
and~\ref{fig:move-cplus} show parts of the robot's domain, in
particular the representation of the action of going to a location,
in SAT (by a set of clauses), in ASP (by a set of rules) and in
\cplus\ (by a set of causal laws), respectively. In these
representations, $y$ ranges over locations $\{L_1,\dots,L_m\}$, and
$b$ ranges over boxes $\{B_1,\dots,B_n\}$. As you can see from these
formulations, it is hard to understand which clauses in the SAT formulation
describe preconditions and which clauses represent direct effects. The ASP
formulation is more concise and it is slightly easier to understand
each rule; however, it is still hard to figure out which kind of rules
(one with nothing on the left-hand-side of the arrow, or one
with some atom) to use for representing what.
The \cplus\ formulation is closer to natural language and easier to understand:
 causal laws of the form
    $\nonexecutableIf{a}{f}$ describe preconditions $\neg f$ of
    an action $a$; causal laws of the form $\causes{a}{f}$ describe
    direct effects $f$ of an action $a$; and causal laws of the form
    $\static{f}{g}$ describe ramifications of actions.
However, we still need to know about the syntax and semantics of formulas in \cplus\ to formalize
a robotic domain.

\begin{figure}[t!]
\hrule

\vspace{0.1cm}

The robot cannot be at two different locations:\\
$\ba l
\neg \ii{atRobo}(x,t) \vee \neg\ii{atRobo}(y,t) \qquad (x<y) \\
\ea$

\medskip
An object $b$ cannot be at two different locations:\\
$\ba l
\neg\ii{atObj}(b,x,t)\vee \neg\ii{atObj}(b,y,t) \qquad (x<y) \\
\ea$

\medskip
Preconditions of $\ii{goto}(y,t)$:\\
$\ba l
     \neg\ii{goto}(y,t)\vee \neg\ii{atRobo}(y,t) \\
     \neg\ii{goto}(y,t)\vee \neg\ii{atObj}(b,y,t)
\ea $

\medskip
Direct effects of $\ii{goto}(y,t)$:\\
$\ba l
     \neg \ii{atRobo}(y,t+1) \vee \ii{goto}(y,t) \vee \ii{atRobo}(y,t+1) \\
     \neg \ii{atRobo}(y,t+1) \vee \ii{goto}(y,t) \vee \ii{atRobo}(y,t) \\
     \ii{atRobo}(y,t+1) \vee \neg \ii{goto}(y,t) \\
     \ii{atRobo}(y,t+1) \vee \neg \ii{atRobo}(y,t+1) \vee \neg \ii{atRobo}(y,t) \\
\ea $

\medskip
Ramifications:\\
 $\ba l

     \neg \ii{atObj}(b,y,t) \vee  \ii{holding}(b,t) \vee \ii{atObj}(b,y,t) \\
     \neg \ii{atObj}(b,y,t) \vee  \ii{holding}(b,t) \vee \ii{atObj}(b,y,t-1) \\
     \neg \ii{atObj}(b,y,t) \vee  \ii{atRobo}(y,t) \vee \ii{atObj}(b,y,t) \\
     \neg \ii{atObj}(b,y,t) \vee  \ii{atRobo}(y,t) \vee \ii{atObj}(b,y,t-1) \\
     \ii{atObj}(b,y,t) \vee  \neg \ii{holding}(b,t) \vee \neg \ii{atObj}(b,y,t) \\
     \ii{atObj}(b,y,t) \vee  \neg \ii{atObj}(b,y,t) \vee \neg \ii{atObj}(b,y,t-1)\\
\dots
     \ea $
     \hrule
  \caption{Describing the robot's domain in SAT}
  \vspace{0\baselineskip}
  \label{fig:move-sat}
  \end{figure}

\begin{figure}[t!]
\hrule

\vspace{0.1cm}

The robot cannot be at two different locations:\\
$\ba l
\ar \ii{atRobo}(x,t), \ii{atRobo}(y,t) \qquad (x<y) \\
\ea$

\medskip
An object $b$ cannot be at two different locations:\\
$\ba l
\ar \ii{atObj}(b,x,t), \ii{atObj}(b,y,t) \qquad (x<y) \\
\ea$

\medskip
Preconditions of $\ii{goto}(y,t)$:\\
$\ba l
     \ar \ii{goto}(y,t), \ii{atRobo}(y,t) \\
     \ar \ii{goto}(y,t), \ii{atObj}(b,y,t)
\ea $

\medskip
Direct effects of $\ii{goto}(y,t)$:\\
$\ba l
     \ii{atRobo}(y,t+1) \ar \ii{goto}(y,t)
\ea $

\medskip
Ramifications:\\
 $\ba l
     \ii{atObj}(b,y,t) \ar \ii{holding}(b,t), \ii{atRobo}(y,t)\\
\dots
     \ea $
     \hrule
  \caption{Describing the robot's domain in ASP}
  \vspace{0\baselineskip}
  \label{fig:move-asp}
  \end{figure}

\begin{figure}[t!]
\hrule

\vspace{0.1cm}

Preconditions of $\ii{goto}(y)$:\\
$\ba l
     \nonexecutableIf{\ii{goto}(y)}{\ii{atRobo}=y} \\
     \nonexecutableIf{\ii{goto}(y)}{\ii{atObj}(o)=y}
\ea $

\medskip

Direct effects of $\ii{goto}(y)$:\\
$\ba l
     \causes{\ii{goto}(y)}{\ii{atRobo}=y}
\ea $

\medskip

Ramifications:\\
 $\ba l
     \static{\ii{atObj}(b)=y}{\ii{holding}(b)\wedge\ii{atRobo}=y}
     \\
\dots
     \ea $
     \hrule
  \caption{Describing the robot's domain in \cplus}
  \vspace{0\baselineskip}
  \label{fig:move-cplus}
  \end{figure}

\begin{figure*}[htb]
      \centering
      \resizebox{4.75in}{!}{\includegraphics{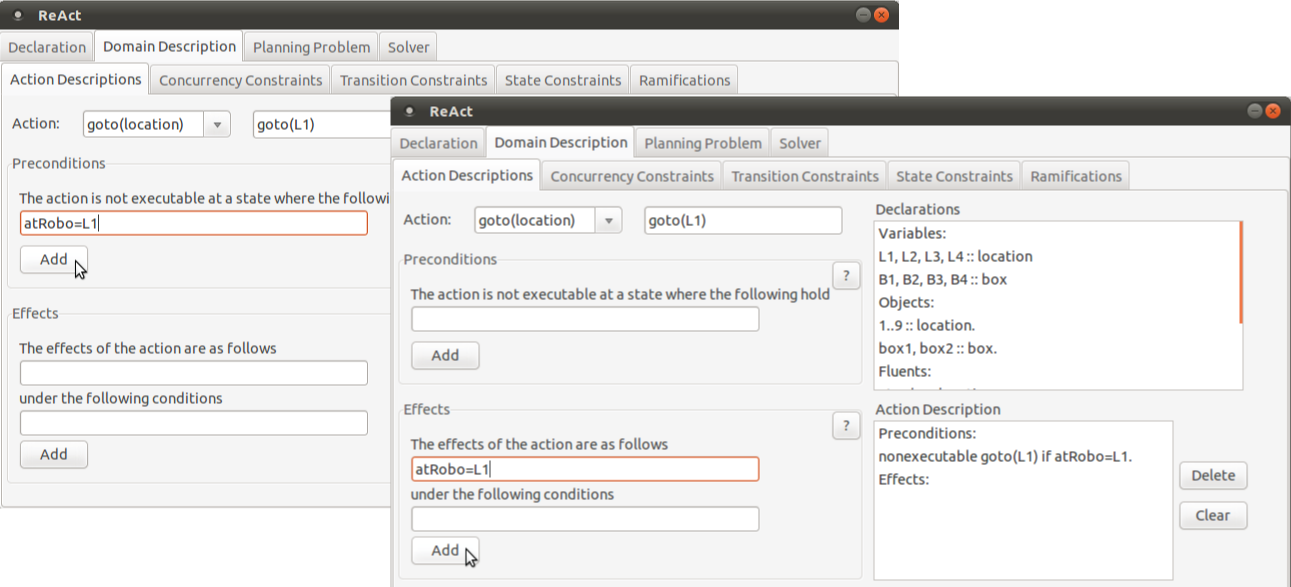}}
      \vspace*{-0.5\baselineskip}
      \caption{A precondition of {\tt goto(L1)} (i.e., the robot is not already at {\tt L1}),
      and a direct effect of {\tt goto(L1)} (i.e., the robot is at {\tt L1}).}
      \label{fig:precondition}
      \vspace*{0\baselineskip}
\end{figure*}

\begin{figure*}[htb]
      \centering
            \resizebox{4.750in}{!}{\includegraphics{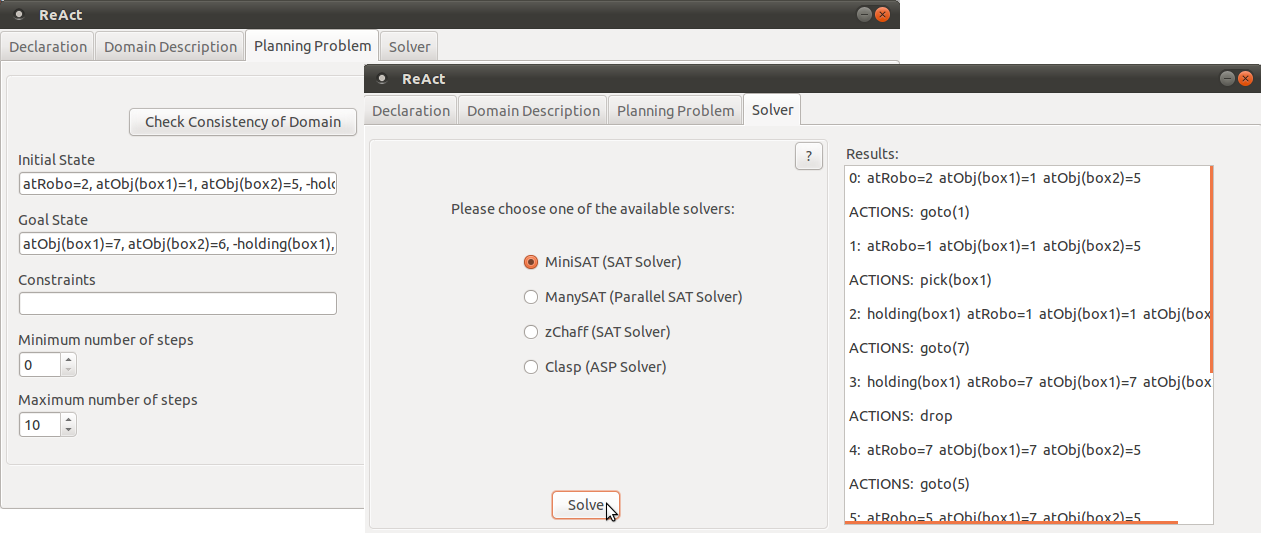}}
      \vspace*{-0.50\baselineskip}
      \caption{A planning problem.}
      \label{fig:pblm}
      \vspace*{0\baselineskip}
\end{figure*}

Moreover, once the domain is represented and the reasoning problem is
specified formally, to solve a given planning problem, we need to know
the specific usage and/or command lines of the relevant automated reasoners.
Therefore, it is challenging and time-consuming to learn how dynamic domains
can be represented in such a logic-based formalism, and how automated reasoners
can be used to solve planning problems (and other reasoning problems, such as prediction and postdiction).

We would like to enable the robotic researchers to start
using automated reasoners to solve various planning problems in
dynamic domains with robotic agents, and assist them to have a better understanding
of the concepts on reasoning about actions and change, so that they
can build/use robots that are furnished with deliberate reasoning capabilities
to autonomously perform some tasks. With this motivation, we have built an interactive tool that guides robotic researchers

\begin{itemize}
\item to represent dynamic
domains in a generic way (so they do not have to know about a
specific action description language),

\item to embed continuous geometric reasoning in discrete task planning in a modular way
 (so they do not have to modify the source codes of relevant planners or implement new hybrid planning algorithms), and

\item to solve planning problems using various
planners/reasoners (so they do not have to know any specifics of
these systems).
\end{itemize}

\section{\react} \label{Sec:react}

\react\ is an interactive tool that helps the users represent a
dynamic domain in a logic-based formalism, and solve a planning
problem in this domain using an automated reasoner. It
guides the users with an interactive user-interface providing
explanations and examples, so that the users do not have to know about
the underlying formalism or the reasoner. \react\ also assists users to
have an understanding of fundamental concepts in knowledge representation and reasoning.

For instance, the preconditions and direct effects of the action of
going to a location are described in \react\ as depicted in
Figure~\ref{fig:precondition}. As seen in the upper inset, the user
describes (in the left part of the user-interface) the following
precondition of $\ii{goto}(L1)$: the robot is not already at $L1$.
While describing the precondition, the user can use the variables
and the object constants (shown on the right part of the
user-interface) declared earlier, with the aid of auto-completion.
After the user adds this precondition, it is displayed in the syntax
of \cplus\ on the right part of the interface, as shown in the lower
inset in Figure~\ref{fig:precondition}. The question mark symbols on
the user-interface provide information about what preconditions and
effects are, how they look like, with examples.

\begin{figure*}[htb]
      \centering
            \resizebox{4.75in}{!}{\includegraphics{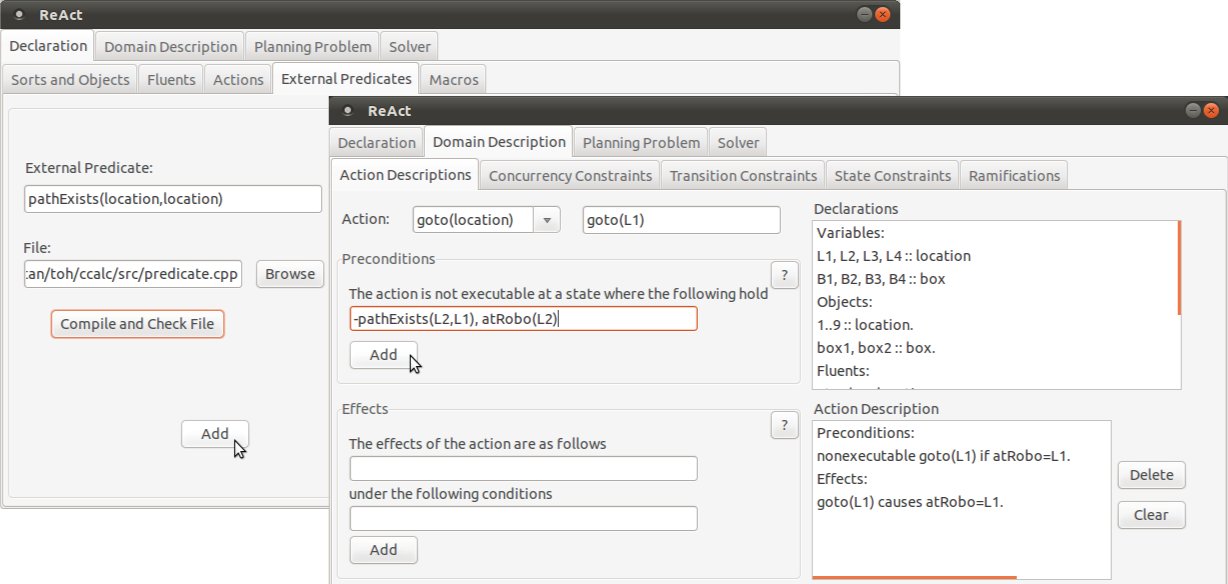}}
      \vspace*{-0.5\baselineskip}
      \caption{Embedding geometric reasoning (i.e., to check existence
      of a collision-free path) as part of a precondition of {\tt goto(L1)}:
      if the robot is at location {\tt L2} then it can goto (L1)
      if there is a collision-free path between {\tt L1} and {\tt L2}.}
      \label{fig:ext}
      \vspace*{0\baselineskip}
\end{figure*}

Once the action domain is described, the user checks whether the
domain description is consistent. If the description is consistent,
then the user continues with the specification of a planning problem
by means of an initial state and a goal state, as shown in the upper
inset in Figure~\ref{fig:pblm}. This problem can then be solved by
an automated reasoner.

Currently, the underlying formalism of \react\ is the action description language \cplus.
The user can choose one of the state-of-the-art SAT solvers among \chaff,
\minisat\ and \manysat, or the state-of-the-art
ASP solver \clasp\ as an automated reasoner.
If the user chooses a SAT solver, \react\ uses
\ccalc~\cite{McCainT97} to transform the action domain description in
\cplus\ to a set of clauses, and calls the SAT solver to find a plan
for the given problem with respect to the described action domain,
as shown in the lower inset in Figure~\ref{fig:pblm}.
If the user chooses an ASP solver, \react\ uses the program
\cplusasp~\cite{CasolaryL11} to transform the action domain description in \cplus\ to a set of
rules in ASP, and calls \clasp\ to find a plan for the given problem
with respect to the described action domain (Figure~\ref{fig:pblm}).

\subsection{Hybrid Planning with \react}

\react\ allows integration of continuous geometric constraints (e.g., by calling a motion planner
to check whether going to some location by means of a continuous
trajectory is feasible without colliding to any static object) as well as
temporal constraints (e.g., by defining durative actions and imposing
deadlines for completion of tasks) into
discrete high-level representation of a domain description, so that
the discrete plan computed by \react\ is feasible. Such an
integration is possible by means of ``external
predicates''~\cite{McCain_thesis,EiterIST06}---predicates that are not part of the action domain
description (i.e., different from fluents and actions) but that are
defined externally in some programming language (e.g., C++, Prolog).
Integrating geometric reasoning (in
preconditions of actions) and temporal reasoning (in planning
problems) are possible thanks to the expressive input language of
\ccalc. Essentially, \ccalc\ performs some preprocessing of external predicates,
while transforming causal laws into propositional logic formulas. The ASP solver \clasp\ also supports external predicates.

\begin{figure*}[htb]
      \centering
      \resizebox{4.75in}{!}{\includegraphics{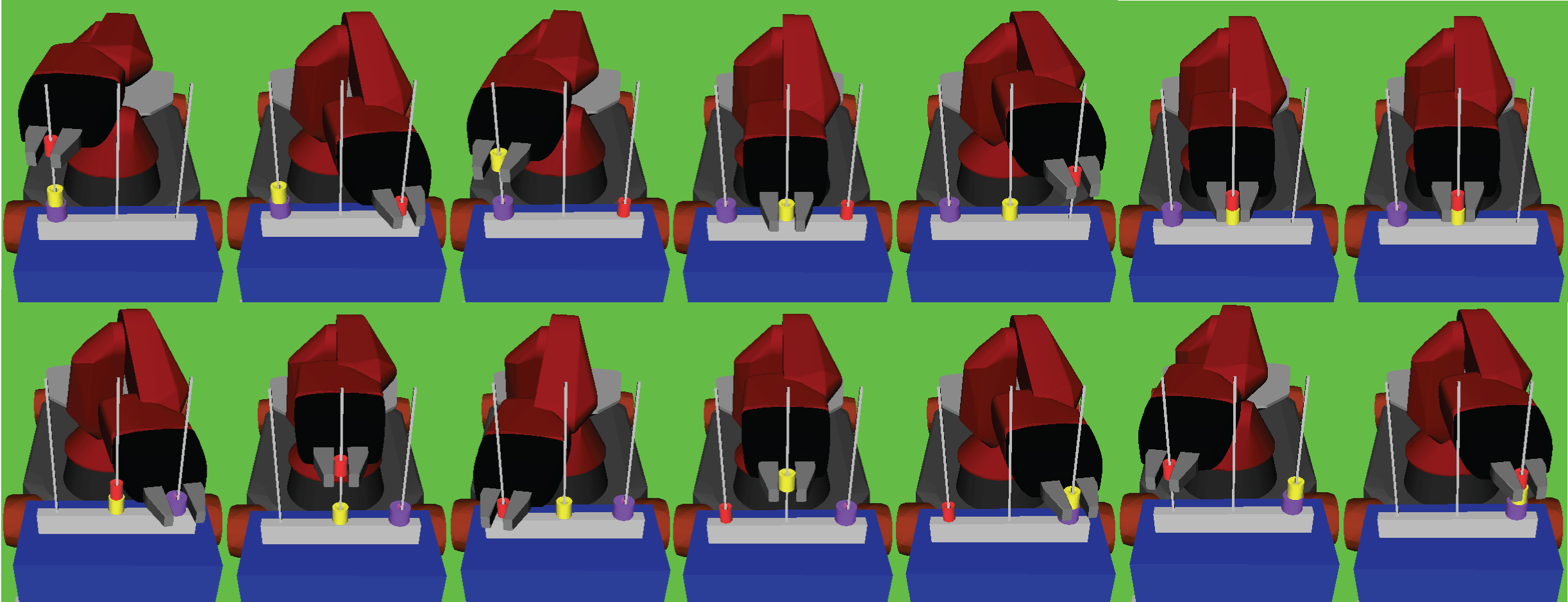}}
      \vspace*{-0.5\baselineskip}
      \caption{\react\ provides dynamic simulations for solutions of Tower of Hanoi problem instances.}
      \label{fig:toh-sim}
      \vspace*{0\baselineskip}
\end{figure*}

The lower inset in  Figure~\ref{fig:ext} shows an example of
integrating collision checking as part of the preconditions of the
action of going to some location $L_1$ (from a location $L_2$), by
means of an external predicate $\ii{pathExists}(L_1,L_2)$ declared
earlier as in the upper inset. By this way, geometric reasoning is
"embedded" in high-level representation: while computing a
task-plan, the reasoner takes into account geometric
models and kinematic relations by means of external predicates
implemented for geometric reasoning. In that sense,
the geometric reasoner guides the  reasoner to find
feasible solutions.

An interesting example of hybrid planning with temporal constraints is given for the
housekeeping domain in~\cite{syroco12}. In this domain, not only an external predicate $\ii{pathExists}$
is used to check geometric feasibility of action of going to some location, but also a second
external predicate $\ii{timeEstimate}$ is utilized to estimate the time it will take for the robot to
traverse the path. Then,  while describing the planning problem, temporal constraints, for instance the constraint
that the total time required to complete the plan should be less than a predefined value, can be added to the specification of a planning problem (the upper inset of Figure~\ref{fig:pblm}).

Hybrid planning with geometric and temporal constraints has been applied in various domains
(e.g., cognitive factories~\cite{etfa12}, service robotics~\cite{syroco12,lpnmr11}, robotic manipulation~\cite{icra11,icra13})
in the spirit of cognitive robotics~\cite{ErdemP12}. \react\ allows formulations of these domains.

\subsection{Sample Domains and Simulation Interface of \react}
To facilitate the use of \react\ for robotic applications and help robotic researchers to
gain experience on fundamental concepts in reasoning about actions and change, \react\
provides a set of example domains, including Tower of Hanoi and Multi-Agent
Path Planning problems.
In each example domain, \react\ provides explanations
at each tab (e.g., about the concepts of a fluent, an action, preconditions/effects of an action, ramifications, static/transition constraints, planning problem), so that the user can have a better understanding of systematically
representing a dynamic domain. 
Once a plan is computed by an automated reasoner, \react\ also provides
a dynamic simulation for an execution of the plan, using \openrave\ ~\cite{diankov_thesis}.
 For the Tower of Hanoi example,
\react\ also provides a wrapper to execute the plans on a physical
KuKa youBot manipulator through
Robot Operating System (ROS). 
For instance, Figure~\ref{fig:toh-sim} shows a snapshot of a
simulation of a plan computed by \minisat\ for a Tower of Hanoi
problem, while a movie of its physical implementation can be viewed
from the following link: {\small \url{http://cogrobo.sabanciuniv.edu/?p=690}}.

\section{Conclusions} \label{Sec:conclusions}

We have introduced \react\ as an interactive tool for cognitive
robotic applications. Significantly reducing the learning time,
\react\ lets robotic researchers to concentrate on robotic
applications, by enabling them to describe action domains
systematically and to solve complex problems relevant to
robotics applications using automated reasoners. \react\ not only
assists its users to have a better understanding
of the concepts on reasoning about actions and change, but also
provides sample formalizations of some action domains, as well as dynamic simulations of plans
computed by a selected automated reasoner. 


\section*{Acknowledgments}

This work is partially supported by TUBITAK 111E116.

\bibliographystyle{splncs}

\label{lastpage}
\end{document}